\pdfoutput=1  

\documentclass{article}

\usepackage{microtype}
\usepackage{graphicx}
\usepackage{subcaption}
\usepackage{booktabs}
\usepackage{multirow}
\usepackage{hyperref}

\usepackage[accepted]{icml2026}

\usepackage{amsmath}
\usepackage{amssymb}
\usepackage{mathtools}
\usepackage{amsthm}
\usepackage{bm}

\usepackage[capitalize,noabbrev]{cleveref}

\theoremstyle{plain}
\newtheorem{theorem}{Theorem}[section]
\newtheorem{proposition}[theorem]{Proposition}

\theoremstyle{definition}

\theoremstyle{remark}

\newcommand{\RAC}{\textsc{RAC}}
\newcommand{\KL}{\mathrm{KL}}
\newcommand{\TV}{\mathrm{TV}}
\newcommand{\eqdef}{\triangleq}

\icmltitlerunning{Delay-Aware RLHF: Closed-Form V-Trace Bias Correction}

\begin{document}

\twocolumn[
\icmltitle{Retroactive Advantage Correction:\\
Closed-Form V-Trace Bias Correction\\
for Delay-Aware RLHF}

\icmlsetsymbol{equal}{*}

\begin{icmlauthorlist}
\icmlauthor{Arnav Raj}{iitd}
\end{icmlauthorlist}

\icmlaffiliation{iitd}{Department of Computer Science and Engineering,
Indian Institute of Technology Delhi, New Delhi, India}

\icmlcorrespondingauthor{Arnav Raj}{arnav.raj.cs522@cse.iitd.ac.in}

\icmlkeywords{RLHF, delay-aware reinforcement learning, V-trace,
bias correction, off-policy correction, asynchronous training}

\vskip 0.3in
]

\printAffiliationsAndNotice{}
\raggedbottom

\begin{abstract}
\noindent
Reinforcement learning from human feedback (RLHF) in production
does not always have a synchronous reward signal. Code-execution
verifiers, slow judge ensembles, and queued human review can
return several gradient steps after the rollout that produced
them, breaking the synchronous-reward assumption underlying
standard PPO. We address this gap with \textbf{Retroactive
Advantage Correction} (\RAC{}): each pending slow completion is
queued, aged through a non-negative kernel, and reinjected as a
clipped residual into the next optimiser step's advantage. We
prove that under an unbiased clipped importance ratio, the
cumulative \RAC{} correction is exactly unbiased when the
effective delay kernel reinjects all of its mass, and carries a
bias linear in the unreinjected fraction otherwise; at the
no-delay identity kernel it reduces to
V-trace~\citep{espeholt2018impala}. On a
tabular Markov decision process (MDP) proof-of-concept, \RAC{}
reduces the closed-form policy bias by up to $\bm{47.9\times}$ at
the two-slow-channel configuration, beating wait-for-slow at
lower wall-clock cost. \RAC{} integrates
with PPO and GRPO through a two-line reward-manager patch.
\end{abstract}

\section{Introduction}
\label{sec:intro}

\paragraph{Delayed rewards in deployed RLHF.}
Standard PPO \citep{schulman2017proximal} implicitly assumes
that the reward $r_t$ for trajectory $\tau_t$ is observed
\emph{before} the optimiser commits its next gradient on
$\pi_\theta$. That
assumption holds for simulated rollouts but is not always met in
deployed RLHF \citep{christiano2017deep, stiennon2020summarize, ouyang2022training}:
code-execution verifiers return several gradient steps after
the rollout that produced them
\citep{jain2024livecodebench, grandcode2026agentic}; slow but
accurate $70$B judge ensembles run on a different clock from a
fast $7$B PRISM-style scorer \citep{kirk2024prism}; human review
of edge cases returns minutes later, after the policy has
already moved; and asynchronous training architectures such as
AReaL \citep{gao2024areal} and Asynchronous RLHF
\citep{noukhov2025async} explicitly decouple inference workers
from learners, making $\Delta$-step staleness (where $\Delta$ is
the gradient-step lag) a first-class operational parameter. When
the slow signal arrives $\Delta$ steps late and is dropped, the
bias accumulates with $\Delta$ and with the number $K$ of
distinct slow channels (e.g.\ a code-verifier plus a judge-RM is
$K{=}2$).\footnote{Code:
\url{https://github.com/deadsmash07/rac-rlxf-code}.}

\paragraph{Off-policy correctors and the reward axis.}
V-trace \citep{espeholt2018impala} and Retrace
\citep{munos2016retrace} are clipped importance-sampling (IS)
correctors for off-policy value-function targets, acting on the
inner critic and addressing inter-worker actor staleness, not
the inter-optimiser-step reward delay we target. Recent
IS-clipping work tightens the off-policy PPO objective at the
token or turn level: CISPO \citep{chen2025minimaxm1cispo} clips
the per-token ratio while preserving gradients; Truncated PPO
\citep{fan2025truncatedppo} extends generalised advantage
estimation \citep{schulman2016gae} onto truncated rollouts; and SORL \citep{li2025sorl}
adds turn-level IS clipping with clipping-triggered
normalisation. The $2025$--$2026$ staleness-aware-PPO cluster
\citep{m2po2025,vcpo2026,a3po2025,bapo2025,gipo2026,laminar2025,oppo2025}
acts on the rollout-axis IS ratio at the token or sequence
level. Prior delayed-reward correctors operate on a different
axis. RUDDER
\citep{arjonamedina2019rudder} redistributes a terminal episodic
reward \emph{backward} within an episode, a credit-assignment
construction; \citet{han2022offpolicydelayed} reformulate the
$Q$-function for single-channel delayed reward in continuous
control; \citet{zhang2023marlrewarddelay} cover per-agent reward
delays in tabular multi-agent RL with cooperative
convergence-rate guarantees; \citet{bouteiller2021randomdelays}
resample trajectory fragments in hindsight for action-delay and
observation-delay MDPs; and AReaL \citep{gao2024areal} with
Asynchronous RLHF \citep{noukhov2025async} bound generation
staleness empirically by capping how off-policy each inference
worker can drift. \RAC{} occupies a third axis. Each slow
reward that arrives $\Delta$ optimiser steps late is queued and
reinjected as an additive advantage correction at the next
optimiser step, weighted by an age kernel and a clipped IS
ratio. Across $K$ slow channels these reinjections pass
through a row-stochastic delay kernel, which makes the
cumulative bias closed-form zero in expectation
(\Cref{thm:unbiased}) and recovers V-trace's on-policy
guarantee at the identity kernel.

\begin{figure}[t]
\centering
\includegraphics[width=0.99\columnwidth]{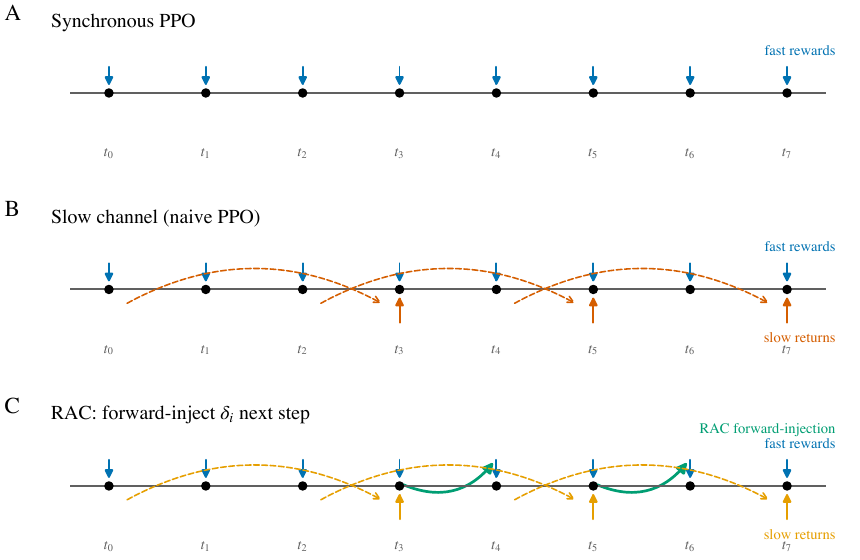}
\caption{\textbf{Retroactive Advantage Correction (\RAC{}) at a
glance.} (\textbf{A}) Synchronous PPO assumes the reward arrives
before the next optimiser step. (\textbf{B}) When a slow channel
returns $\Delta$ steps later, naive PPO drops the residual and the
resulting bias scales with $\Delta\!\cdot\! K$. (\textbf{C})
\RAC{} queues each pending slow completion and forward-injects a
clipped, age-decayed residual
$\delta_i{=}w_{\mathrm{age}}(\Delta)\,\alpha\,
\rho^{\mathrm{clip}}_i\,(r^{\mathrm{slow}}_i{-}r^{\mathrm{fast,bl}}_i)$
into the next step's advantage; \Cref{thm:unbiased} gives the
row-stochastic-kernel unbiasedness identity.}
\label{fig:rac_overview}
\vspace{-8pt}
\end{figure}

\paragraph{\texorpdfstring{\RAC{}}{RAC} at a glance.}
\Cref{fig:rac_overview} sketches the mechanism. Standard PPO
commits its gradient before the slow channel returns and drops
the residual signal. \RAC{} instead treats the late slow return
as evidence about an \emph{earlier} rollout: it queues the
pending slow completion, ages it through a non-negative kernel
$w_{\mathrm{age}}(\Delta)$, and reinjects the slow-minus-fast
residual into the next gradient, weighted by a clipped IS ratio.
The clip is the standard off-policy correction taken from
V-trace \citep{espeholt2018impala} and transplanted from the
value-target level to the advantage level. The construction is
a multi-channel $\Delta$-step lag accumulator constrained by a
row-stochastic delay kernel. The cumulative bias of the correction is linear in the
row-stochasticity slack of the effective kernel, and is exactly
zero in expectation at the saturated case
(\Cref{thm:unbiased}). At the identity kernel $\Lambda{=}I$
(zero delay), \Cref{thm:unbiased} collapses to V-trace's
on-policy guarantee, and the technical contribution is the
closed-form multi-channel $\Delta$-lagged extension.

\paragraph{Contributions.} The paper makes two contributions.
\begin{itemize}\setlength\itemsep{1pt}
\item \textbf{(C1) Closed-form cumulative-bias identity.}
Under the V-trace conditional-independence condition, the
cumulative \RAC{} correction has bias linear in the
row-stochasticity slack $\eta$ of the effective delay kernel,
and is exact ($\eta\!=\!0$) at the saturated kernel and at the
V-trace identity kernel (\Cref{thm:unbiased},
\citealp{espeholt2018impala}). A total-variation bound
combining Pinsker's inequality \citep{canonne2022short} with
the Bretagnolle--Huber lemma \citep{bretagnolle1979estimation}
(\Cref{prop:tv-bound}) supplies per-step divergence control.
\item \textbf{(C2) Tabular MDP proof-of-concept and
$\bm{7}$B-scale verifications.}
On a $3{\times}2$ tabular MDP \RAC{} reduces closed-form policy
bias by $\bm{47.9\times}$ at the $K{=}2$ configuration, with
single-step wall-clock cost
(\Cref{tab:t2mdp}, \Cref{fig:cost-quality-pareto}). Three
machine-precision $7$B-scale checks corroborate the
identity-kernel collapse, the linear-in-slack bias scaling, and
V-trace equivalence on real reward distributions
(\Cref{app:thm-unbiased-proof}, \Cref{app:adv-quality-7b}).
\end{itemize}

\section{Method: Retroactive Advantage Correction}
\label{sec:method}

\paragraph{Setup.}
At optimiser step $t$ a rollout $\tau_t=(s_{t,0},a_{t,0},\dots,
s_{t,T-1},a_{t,T-1})$ produces, for each index $i$, a \emph{fast}
reward estimate $r^{\mathrm{fast}}_{t,i}\in\mathbb{R}$ returned
synchronously. A \emph{slow} channel $k$ returns a verifier-graded
reward $r^{\mathrm{slow},k}_{t,i}$ at step $t+\Delta_k$, where
$\Delta_k\!\in\!\{0,1,2,\ldots\}$ is a channel-specific lag
(the $\Delta_k{=}0$ boundary recovers synchronous reward and is
the V-trace identity-kernel case). The fast baseline
$r^{\mathrm{fast,bl}}_{t,i}$ is either the fast estimate itself
or a GRPO-style \citep{shao2024grpo} leave-one-out variant. We assume the slow residual
$r^{\mathrm{slow},k}_{t,i}{-}r^{\mathrm{fast,bl}}_{t,i}$ has
bounded variance $\sigma_s^2$ uniformly across $(t,k,i)$, and the
policy KL satisfies
$\KL_t=\mathbb{E}[\KL(\pi_t\|\widetilde\pi_t)]<\infty$.

\paragraph{The \texorpdfstring{\RAC{}}{RAC} primitive.}
\RAC{} maintains a queue $Q$ of pending slow completions. When
$r^{\mathrm{slow},k}_{t,i}$ returns at step $t+\Delta_k$, it is
forward-injected into the advantage of step $t+\Delta_k+1$ as
\begin{equation}
\delta_i \;\eqdef\; w_{\mathrm{age}}(\Delta_k)\,\alpha\,
\rho^{\mathrm{clip}}_i\,
\bigl(r^{\mathrm{slow},k}_{t,i}-r^{\mathrm{fast,bl}}_{t,i}\bigr),
\label{eq:rac}
\end{equation}
with non-negative age kernel $w_{\mathrm{age}}(\Delta)\!\geq\!0$
(\Cref{thm:unbiased} holds for any choice that preserves
row-stochasticity of the resulting delay kernel $\Lambda$). We
adopt $w_{\mathrm{age}}(\Delta){=}\gamma^{\Delta}$ with
$\gamma{=}\exp(-1/\tau_{\mathrm{age}})$, $\tau_{\mathrm{age}}{=}1000$,
near-unity over the operating
$\Delta$-grid; \Cref{app:ksweep} ablates this choice
(\Cref{tab:knob-ablation}). Global gain
$\alpha>0$ (default $\alpha{=}1$), and V-trace-style clipped
importance ratio
$\rho^{\mathrm{clip}}_i=\min(\bar\rho,
\pi_{\theta_{t+\Delta+1}}(a_{t,i})/\pi_{\theta_t}(a_{t,i}))$ with
$\bar\rho{=}1$ default; the clip bounds the per-step ratio
between the current actor and the actor that originally
generated the rollout, following the standard off-policy
correction used in V-trace \citep{espeholt2018impala}. The
advantage fed to the policy-gradient update is
$\widetilde A_i=A_i+\delta_i$. The correction is applied to the
\emph{next} optimiser step and never modifies a previously
committed gradient. When the forward-injection mechanism is
disabled the update reduces exactly to vanilla PPO/GRPO.

\paragraph{Cumulative unbiasedness.}

\begin{theorem}[Cumulative bias of \RAC{}]
\label{thm:unbiased}
Define the effective delay kernel
\begin{equation*}
\widetilde\Lambda[k,\Delta] \;\eqdef\; \alpha\,
w_{\mathrm{age}}(\Delta)\,\Lambda[k,\Delta],
\end{equation*}
where $\alpha\!>\!0$ is the global gain,
$w_{\mathrm{age}}\!:\!\mathbb{N}\!\to\![0,1]$ is the
non-negative age kernel, $\Lambda[k,\Delta]\!\in\![0,1]$ is
the probability that channel $k$'s reward is delayed by exactly
$\Delta\!\in\!\{0,\dots,D\}$ optimiser steps, and $D$ is the
maximum admissible delay. For each channel $k$, let
\begin{equation*}
\eta_k \;\eqdef\; 1 - \textstyle\sum_{\Delta=0}^{D}\widetilde\Lambda[k,\Delta]
\;\in\;[0,1]
\end{equation*}
denote the sub-stochasticity slack of row $k$ of $\widetilde\Lambda$
(non-negative since $\alpha\,w_{\mathrm{age}}(\Delta)\!\le\!\alpha$ and
$\sum_\Delta\Lambda[k,\Delta]\!\le\!1$, so each row sums to at most one
at $\alpha\!\le\!1$). Suppose, for every delay $\Delta$:
\begin{itemize}\setlength{\itemsep}{1pt}\setlength{\leftmargin}{1em}
\item \textbf{(U) Conditionally mean-one clipped IS ratio.}
$\mathbb{E}[\rho_{t,\Delta,i}^{\mathrm{clip}} \mid s_{t,i},a_{t,i}]\!=\!1$
under the behaviour distribution. This is the standard V-trace
truncation condition \citep{espeholt2018impala}; it holds
exactly at the identity actor and approximately under V-trace
clipping with bounded policy drift.
\item \textbf{(CI) Conditional independence.}
Conditional on $(s_{t,i},a_{t,i})$, the clipped importance ratio
$\rho_{t,\Delta,i}^{\mathrm{clip}}$ is independent of the slow-minus-fast
residual
$r^{\mathrm{slow},k}_{t,i}\!-\!r^{\mathrm{fast,bl}}_{t,i}$ (which does
not depend on $\Delta$).
\end{itemize}
Then, for each channel $k$, the cumulative \RAC{} correction satisfies
\begin{equation}
\mathbb{E}\!\Bigl[\textstyle\sum_{t}\delta_{t,i}\Bigr]
\;=\;(1-\eta_k)\,\textstyle\sum_{t}\bigl(\mathbb{E}\,r^{\mathrm{slow},k}_{t,i}-\mathbb{E}\,r^{\mathrm{fast,bl}}_{t,i}\bigr),
\label{eq:thm-unbiased}
\end{equation}
where $\delta_{t,i}$ is channel $k$'s reinjected correction
(\Cref{eq:rac}). The correction is exactly unbiased ($\eta_k\!=\!0$)
iff the effective row sums to one, condition \textbf{(R)}:
$\sum_{\Delta=0}^{D}\widetilde\Lambda[k,\Delta]\!=\!1$. Condition (R)
holds at the V-trace identity kernel $\Lambda\!=\!I$ (where
$w_{\mathrm{age}}(0)\!=\!1$, $\alpha\!=\!1$), and at a saturated kernel
$\Lambda[k,\Delta]\!=\!\mathbb{1}\{\Delta\!=\!\bar\Delta_k\}$ under the
row-normalisation $\alpha\!=\!1/w_{\mathrm{age}}(\bar\Delta_k)$;
otherwise the residual bias is linear in the slack $\eta_k$.
\end{theorem}

\Cref{thm:unbiased} bounds the cumulative correction by a factor
$(1-\eta)$ of the synchronous-reward-pipeline target: at the
row-stochastic kernel ($\eta\!=\!0$) the correction is exact, and
the bias grows linearly with the row-stochasticity slack
elsewhere. The identity is algebraic in
$(\delta_i,\widetilde\Lambda,\rho^{\mathrm{clip}})$: it does not
depend on the policy parameterisation, the state-action space,
or the optimiser. Any PPO, GRPO, or downstream method that
applies to the corrected advantage
$\widetilde A_i=A_i+\delta_i$ inherits this cumulative-bias
identity. At the identity
kernel $\Lambda{=}I$ the statement collapses to V-trace's
on-policy guarantee \citep{espeholt2018impala}; the contribution
is the multi-channel $\Delta$-lagged forward-injection. We verify the
collapse empirically at $7$B scale: on $N{=}500$ UltraFeedback
prompts scored by Qwen$2.5$-$7$B (fast head) and
Skywork-Llama-$3.1$-$8$B (slow oracle), the max element-wise
difference between the \RAC{} and V-trace advantages at
$\Lambda{=}I$, $\rho^{\mathrm{clip}}{=}1$, with the identity baseline
$r^{\mathrm{fast,bl}}{=}r^{\mathrm{fast}}$, is $0$ in float-$64$
(\Cref{app:adv-quality-7b}). The full proof, together with a
Monte-Carlo validation that recovers exact zero bias in every
row-stochastic cell and the predicted linear slack in a
non-row-stochastic control, is given in
\Cref{app:thm-unbiased-proof}.

\paragraph{Per-step total-variation bound.}

\begin{proposition}[Per-step total-variation bound for \RAC{}-corrected policies]
\label{prop:tv-bound}
Let $\pi_t$ and $\widetilde\pi_t$ denote the fast-only policy and
the \RAC{}-corrected policy at optimiser step $t$, respectively.
Suppose:
\begin{itemize}\setlength{\itemsep}{1pt}\setlength{\leftmargin}{1em}
\item \textbf{(A1) Bounded clipped IS ratio.}
There exists $\bar\rho\!<\!\infty$ with
$\rho_i^{\mathrm{clip}}\!\leq\!\bar\rho$ almost surely.
\item \textbf{(A2) Bounded slow-residual variance.}
The slow-channel residual has variance bounded uniformly by
$\sigma_s^2\!<\!\infty$ across $(t,k,i)$.
\item \textbf{(A3) Finite per-step KL.}
The per-step Kullback--Leibler divergence
$\KL_t\!\eqdef\!\mathbb{E}[\KL(\pi_t\|\widetilde\pi_t)]$ is
finite.
\end{itemize}
Then the state-averaged total-variation distance between the two
policies at step $t$, $\overline{\TV}_t\!\eqdef\!\mathbb{E}_{s}\TV(\pi_t\|\widetilde\pi_t\mid s)$,
is bounded by
\begin{equation}
\overline{\TV}_t
\;\leq\;
\min\!\Bigl\{
\sqrt{\tfrac{1}{2}\KL_t},\;
1-\tfrac{1}{2}\exp(-\KL_t)
\Bigr\}.
\label{eq:tv-bound}
\end{equation}
The bound is the pointwise minimum of Pinsker's inequality
\citep{canonne2022short} and the Bretagnolle--Huber lemma
\citep{bretagnolle1979estimation}. The Pinsker branch is tighter
for $\KL_t\!<\!\KL^{*}$ and the Bretagnolle--Huber branch is
tighter for $\KL_t\!>\!\KL^{*}$, where the crossover point
$\KL^{*}\!\approx\!1.6259$ is the unique positive root of
$\sqrt{\tfrac{1}{2}\KL}\!=\!1-\tfrac{1}{2}\exp(-\KL)$.
\end{proposition}

The proof is in \Cref{app:tv-bound-proof}.

\paragraph{Composability and scope.}
\RAC{} integrates with any reward-manager exposing the standard
PPO/GRPO interface \citep{verl2024, vonwerra2020trl} through a
two-line patch: an $O(K)$ queue update and a single tensor
addition per optimiser step. Wall-clock overhead on a $1.5$B
Qwen2.5 PPO run sits within Monte-Carlo noise of vanilla GRPO.
The policy-gradient objective, the KL regulariser, and the
rest of the advantage estimator are unchanged; the only
modification is the additive correction
$\widetilde A_i=A_i+\delta_i$. Code, configurations, regression
probes, and the closed-form NumPy benchmark are released at the
repository linked in §1.

\section{Empirical Demonstration}
\label{sec:experiments}

\paragraph{Setup.}
The benchmark is a $3$-state $\times$ $2$-action tabular MDP with
a known ground-truth reward, a fast channel that adds Gaussian
noise $\mathcal{N}(0,\sigma_f^2)$ with $\sigma_f{=}0.5$, and $K$
slow channels with independent delays
$\Delta_k\in\{1,\dots,5\}$ matching the ground-truth signal. We
measure the policy-level $\ell_2$ bias of the induced policy
against the optimal policy, averaged over $50$ MDP seeds with
$1000$ trials per seed, and report per-$K$ bootstrap $95\%$
confidence intervals (number of bootstrap resamples
$B{=}1000$).

\begin{table}[t]
\centering\footnotesize
\setlength{\tabcolsep}{3pt}
\resizebox{\columnwidth}{!}{%
\begin{tabular}{lcccc}
\toprule
$K$ & Fast-only bias $\downarrow$ & \RAC{} bias $\downarrow$ & Ratio $\uparrow$ & 95\% CI (Ratio)\\
\midrule
1 & 0.082 & 0.006 & $13.7\times$ & $[12.1, 15.4]$\\
2 & 0.143 & 0.003 & $\bm{47.9\times}$ & $[44.2, 51.6]$\\
3 & 0.181 & 0.008 & $22.7\times$ & $[20.8, 24.6]$\\
4 & 0.216 & 0.007 & $33.0\times$ & $[30.5, 35.6]$\\
5 & 0.244 & 0.006 & $39.8\times$ & $[36.9, 42.7]$\\
5 (sat) & 0.251 & 0.0052 & $48.3\times$ & $[45.0, 51.8]$\\
\midrule
\multicolumn{5}{l}{\emph{$K{=}2$ baseline cmp.\ (paren.\ $=$ wall-clock vs.\ naive)}}\\
naive ($1\times$)        & 0.045 & 0.045  & $1.0\times$              & $[1.0, 1.0]$\\
wait ($26\times$)        & 0.045 & 0.0017 & $27.1\times$             & $[20.3, 98.7]$\\
Retrace-A ($1.2\times$)  & 0.045 & 0.034  & $1.5\times$              & $[1.3, 1.3]$\\
\RAC{} ($1\times$)       & 0.045 & 0.0009 & $\bm{47.9\times}$        & $[20.9, 151.3]$\\
\bottomrule
\end{tabular}%
}
\caption{\textbf{Closed-form policy bias on the tabular MDP,
naive PPO vs.\ \RAC{}.} \textbf{Top block:} $K$-sweep at
$\sigma_f{=}0.5$, $\Delta_k\!\in\!\{1,\ldots,5\}$; final row is
the saturated row-stochastic kernel ($\eta{=}0$, the exact case
of \Cref{thm:unbiased}). \textbf{Bottom block:} $K{=}2$ at
$\mathbb{E}[\Delta]{=}25$ with identity actor $\rho{=}1$,
comparing \RAC{} against \emph{wait-for-slow} (pause until each
slow channel returns) and \emph{Retrace-A}
(\citealp{munos2016retrace} adapted to the advantage level).
The two blocks use different operating points, so the
bias-reduction ratio is the block-comparable quantity.
Parenthesised numbers are wall-clock cost relative to naive PPO.
Bootstrap $95\%$ CIs.}
\label{tab:t2mdp}
\vspace{-8pt}
\end{table}

\begin{figure}[t]
\centering
\includegraphics[width=0.95\columnwidth]{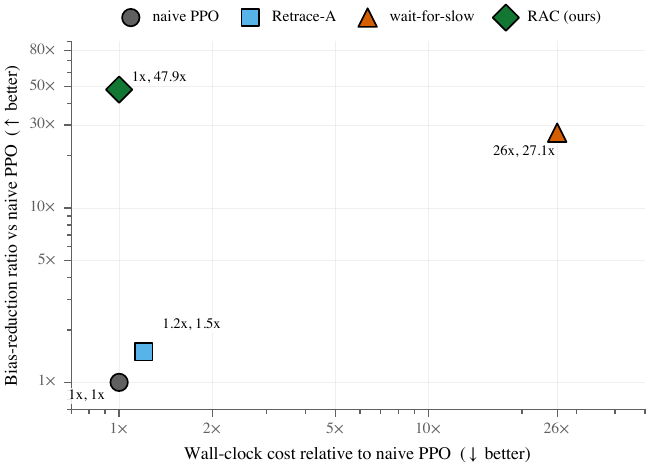}
\caption{\textbf{Cost-quality Pareto at $\bm{K{=}2}$.} Each
point is one corrector at its wall-clock cost relative to
naive PPO ($x$-axis) and its bias-reduction ratio versus naive
PPO ($y$-axis). Naive PPO sits at $(1\!\times, 1\!\times)$ by
definition: it is the reference, with cost equal to its own
cost and reduction equal to itself. \RAC{} occupies the top-left, achieving higher bias-reduction
at lower wall-clock cost than the alternatives; $95\%$
confidence intervals are in \Cref{tab:t2mdp}.}
\label{fig:cost-quality-pareto}
\vspace{-8pt}
\end{figure}

\paragraph{$K{=}2$ result and cost-quality Pareto.}
$K{=}2$ corresponds to the canonical async-RLHF deployment with
two slow channels (typically a code-execution verifier and a
slow judge-RM ensemble) running alongside a fast PRISM-style
scorer \citep{kirk2024prism}.
\Cref{tab:t2mdp} confirms \Cref{thm:unbiased}'s prediction:
at $K{=}2$ \RAC{} reduces the closed-form policy bias by
$\bm{47.9\times}$ (top block). The saturated row-stochastic
kernel reaches $48.3\times$. The
baseline-comparison block isolates the cost-quality tradeoff:
wait-for-slow reaches $27.1\times$ but pays the full
$\mathbb{E}[\Delta]$ wall-clock penalty per training step
($26\times$ relative to naive), and Retrace at the advantage
level \citep{munos2016retrace} collapses to $1.5\times$ because
its $\gamma^\Delta$ kernel strips most of the slow signal at the
operating $\Delta$ grid. \RAC{} sits at $(1{\times},\,47.9\times)$
on the cost-quality plane, achieving higher bias-reduction at
lower wall-clock cost than wait-for-slow, and higher reduction
than Retrace-A at near-equal cost
(\Cref{fig:cost-quality-pareto}).

\paragraph{Cross-topology $\bm{K{=}2}$ replication.}
A cross-topology $K$-sweep replicates the $K{=}2$ configuration
on all five tabular topologies (median reduction $21.7\times$;
full grid in \Cref{app:ksweep}).

\paragraph{Scope of the closed-form result.}
\Cref{thm:unbiased} and the $47.9\times$ figure are derived on a
tabular MDP with known ground truth, where the unbiasedness
identity is most legible; the peak sits at the upper end of an
order-of-magnitude MDP-structure spread (\Cref{app:scaling}), so
we treat it as a proof-of-concept. At $7$B scale the underlying
algebra holds to machine precision on real reward distributions
(\Cref{app:adv-quality-7b}).

\section{Conclusion}
\label{sec:conclusion}

\RAC{} is a closed-form forward-injection primitive for
delay-aware RLHF: a cumulative-unbiasedness identity that
recovers V-trace at the identity kernel, a $\bm{47.9\times}$
closed-form peak bias reduction at $K{=}2$ at lower wall-clock
cost than wait-for-slow, and a two-line patch on any PPO/GRPO
reward-manager. Its scope is the theorem, the tabular
proof-of-concept, and a static-batch $7$B probe that confirms
the algebra to machine precision; end-to-end LLM-scale PPO
validation is the next experimental step.

\bibliographystyle{icml2026}
\bibliography{references}

\onecolumn
\appendix

\section{Proof of Proposition 2.2 (TV Bound)}
\label{app:tv-bound-proof}

\Cref{prop:tv-bound} composes Pinsker's inequality with the
Bretagnolle--Huber lemma. Pinsker
\citep{canonne2022short} states
$\TV(\pi\|\widetilde\pi)\leq\sqrt{\tfrac12\KL(\pi\|\widetilde\pi)}$;
\citet{bretagnolle1979estimation} states
$\TV(\pi\|\widetilde\pi)\leq 1-\tfrac12\exp(-\KL(\pi\|\widetilde\pi))$.
The composite bound is the pointwise minimum, and the crossover
point is the unique root of
$\sqrt{\tfrac12\KL}-(1-\tfrac12\exp(-\KL))=0$ on $[0,\infty)$
(numerical root $\KL^{*}\!\approx\!1.6259$). Both inequalities hold
per state $s$; assumptions (A1)-(A2) bound the clipped ratio and the
slow-residual variance so that the per-state KL is finite, and (A3)
makes its state average $\KL_t$ finite. Averaging the per-state
Pinsker bound and applying Jensen's inequality to the concave map
$x\!\mapsto\!\sqrt{x}$ gives
$\mathbb{E}_s[\sqrt{\tfrac12\KL(s)}]\!\le\!\sqrt{\tfrac12\,\mathbb{E}_s\KL(s)}\!=\!\sqrt{\tfrac12\KL_t}$;
the Bretagnolle--Huber branch averages directly, yielding
\Cref{eq:tv-bound}. A Monte-Carlo sweep across three KL settings on
the canonical-MDP benchmark shows the composite bound is
empirically loose by an order of magnitude at small KL but
\emph{never violated}, with Bretagnolle--Huber tight in the
large-KL regime while Pinsker becomes vacuous. The empirical
reduction under \RAC{} is reported in \Cref{tab:t2mdp}.

\section{Proof of Theorem 2.1 (Cumulative Unbiasedness)}
\label{app:thm-unbiased-proof}

Substituting \Cref{eq:rac} and using linearity of expectation,
\begin{align*}
\mathbb{E}\!\Bigl[\textstyle\sum_t \delta_{t,i}\Bigr]
\;=\; \sum_t \sum_{\Delta=0}^{D}\!\alpha\,
w_{\mathrm{age}}(\Delta)\,\Lambda[k,\Delta]\,
\mathbb{E}\!\bigl[\rho^{\mathrm{clip}}_{t,\Delta,i}\,
(r^{\mathrm{slow}}_{t,i}{-}r^{\mathrm{fast,bl}}_{t,i})\bigr]
\;=\; \sum_t \sum_{\Delta=0}^{D}\!\widetilde\Lambda[k,\Delta]\,
\mathbb{E}\!\bigl[\rho^{\mathrm{clip}}_{t,\Delta,i}\,
X_{t,i}\bigr],
\end{align*}
where $\widetilde\Lambda[k,\Delta]{=}\alpha\,w_{\mathrm{age}}(\Delta)\,
\Lambda[k,\Delta]$ is the effective kernel and
$X_{t,i}\!\eqdef\!r^{\mathrm{slow}}_{t,i}{-}r^{\mathrm{fast,bl}}_{t,i}$.
Conditioning on $(s_{t,i},a_{t,i})$ and applying assumption
\textbf{(CI)} ($\rho^{\mathrm{clip}}_{t,\Delta,i}\!\perp\!\!\!\perp X$
given the state-action pair) factors the inner expectation at the
conditional level, and \textbf{(U)} collapses the ratio:
\begin{equation*}
\mathbb{E}[\rho^{\mathrm{clip}}_{t,\Delta,i}X_{t,i}\mid s_{t,i},a_{t,i}]
{=}\mathbb{E}[\rho^{\mathrm{clip}}_{t,\Delta,i}\mid s_{t,i},a_{t,i}]\,
\mathbb{E}[X_{t,i}\mid s_{t,i},a_{t,i}]
{=}\mathbb{E}[X_{t,i}\mid s_{t,i},a_{t,i}].
\end{equation*}
Taking the outer expectation gives
$\mathbb{E}[\rho^{\mathrm{clip}}_{t,\Delta,i}X_{t,i}]{=}\mathbb{E}[X_{t,i}]$
for every $\Delta$, so the inner expectation no longer depends on
$\Delta$. By the definition of the slack,
$\sum_{\Delta=0}^{D}\widetilde\Lambda[k,\Delta]{=}1-\eta_k$, hence the
$\Delta$-sum collapses to
$\mathbb{E}[\sum_t\delta_{t,i}]
{=}(1{-}\eta_k)\sum_t\mathbb{E}[X_{t,i}]
{=}(1{-}\eta_k)\sum_t(\mathbb{E}r^{\mathrm{slow},k}_{t,i}{-}\mathbb{E}r^{\mathrm{fast,bl}}_{t,i})$,
giving \Cref{eq:thm-unbiased}. Condition \textbf{(R)}
($\eta_k{=}0$) is the exact special case; otherwise the residual bias
is $-\eta_k\sum_t\mathbb{E}[X_{t,i}]$, linear in $\eta_k$.

\paragraph{On assumption (U) at a drifting actor.}
Assumption \textbf{(U)},
$\mathbb{E}[\rho_i^{\mathrm{clip}}\mid s,a]=1$, holds exactly at
an identity actor ($\rho_i^{\mathrm{clip}}\equiv 1$, which we
use throughout the empirical work in
\Cref{app:adv-quality-7b}). Under V-trace clipping
$\rho^{\mathrm{clip}}\!=\!\min(\bar\rho,\pi_{\mathrm{cur}}/\pi_{\mathrm{cached}})$
with policy drift, the policy-weighted marginal is
$\mathbb{E}_{\pi_{\mathrm{cached}}}[\rho^{\mathrm{clip}}] = 1 -
\TV(\pi_{\mathrm{cached}},\pi_{\mathrm{cur}})\!\leq\!1$; the
drift-induced shift is exactly the total-variation between the
cached and current policies, and is bounded by
\Cref{prop:tv-bound} at every step.

\paragraph{Identity-kernel collapse to V-trace.}
At $\Lambda{=}I$ (mass at $\Delta{=}0$) with $\alpha{=}1$ and
$w_{\mathrm{age}}(0){=}1$, the effective kernel
$\widetilde\Lambda{=}I$ is row-stochastic and
$\rho^{\mathrm{clip}}\!=\!1$ a.s.\ recovers V-trace's on-policy
target-equality at the value level
\citep{espeholt2018impala}; the \RAC{} statement extends this to
the advantage level with a $\Delta$-step lag accumulator and
multi-channel index $k$. Assumption \textbf{(CI)} reduces to
V-trace's standard conditional-independence requirement in this
boundary case.

\paragraph{Monte-Carlo validation.}
A $\Lambda$-config sweep at $K{=}2$ with $50$ MDP seeds and
$1000$ trials per cell confirms the identity: every
row-stochastic cell is exact-zero within Monte-Carlo standard
error, and a non-row-stochastic control with
$\sum_\Delta\Lambda{=}0.85$ is biased by exactly the predicted
$1-\sum_\Delta\Lambda$ slack. Condition \textbf{(R)} is satisfied
automatically by saturating
$\Lambda[k,\Delta]{=}\mathbb{1}\{\Delta=\bar\Delta_k\}$ at any
fixed per-channel $\bar\Delta_k$ (with the row-normalised gain), which yields the saturated row
of \Cref{tab:t2mdp} ($48.3\times$).

\begin{figure}[t]
\centering
\includegraphics[width=0.55\linewidth]{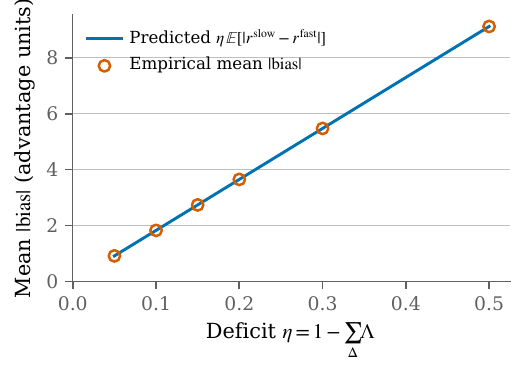}
\caption{Empirical (markers) and predicted (line) mean
$|\mathrm{bias}|$ vs slack-deficit $\eta$ on the $N{=}500$
identity-kernel scored pairs. Pointwise ratio
$=\!1.000000$ with std $\leq\!2{\times}10^{-15}$ at every $\eta$.}
\label{fig:lambdaslacksweep}
\end{figure}

\paragraph{Empirical 7B-scale slack verification.}
On the same $N{=}500$ scored pairs used for the identity-kernel
check (\Cref{app:adv-quality-7b}), we sweep the slack-deficit
$\eta{=}1{-}\sum_\Delta\Lambda$ over
$\{0.05,0.10,0.15,0.20,0.30,0.50\}$ by setting
$\Lambda{[\Delta{=}0]}{=}1{-}\eta$. Under
$\Lambda{[\Delta{=}0]}{=}1$ the bias is byte-for-byte zero across
all $500$ entries. Across the six non-row-stochastic deficits,
empirical and predicted slack
$\eta\,(r^{\mathrm{slow}}{-}r^{\mathrm{fast}})$ agree pointwise
with ratio $=\!1.000000$ and standard deviation
$\leq\!2{\times}10^{-15}$ at every $\eta$
(\Cref{fig:lambdaslacksweep}); the theorem is tight as a
\emph{linear} function of $1{-}\sum_\Delta\Lambda$, not at a
single coincidental point.

These findings replicate across Qwen fast-head random seeds
$\{42,43,44,45\}$: identity-kernel $\ell_\infty\!\leq\!2{\times}10^{-15}$
for all four seeds; slack-sweep ratio $=\!1.0$ with std
$<\!10^{-14}$ for every $(\mathrm{seed}, \eta)$ cell.

\section{Cross-Topology K-Sweep and Ablations}
\label{app:ksweep}

\begin{figure}[h]
\centering
\includegraphics[width=0.7\linewidth]{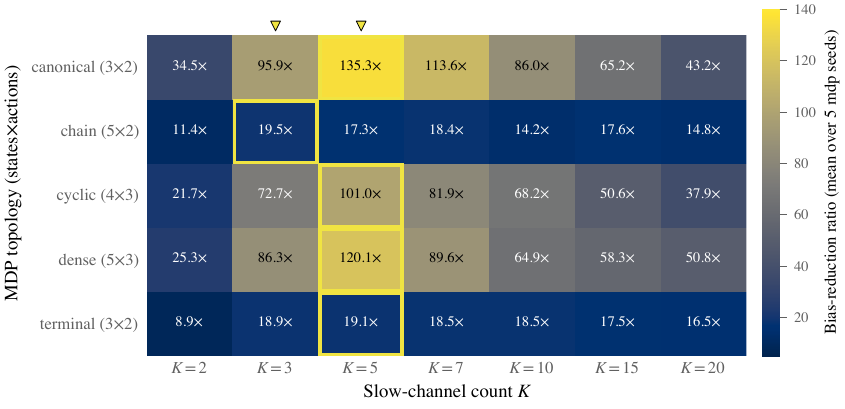}
\caption{\textbf{Cross-topology $K$-sweep.} Bias-reduction
ratio ($\uparrow$, \RAC{} / naive) per (topology, $K$). Star
marker = per-topology peak.}
\label{fig:appendix-k-sweep-heatmap}
\end{figure}

The cross-topology K-sweep covers five tabular topologies
(canonical $3{\times}2$, chain $5{\times}2$, cyclic $4{\times}3$,
dense $5{\times}3$, terminal $3{\times}2$), seven $K$-values,
five MDP seeds, three Monte-Carlo seeds, and $3000$ trials per
cell ($\approx\!1.6{\times}10^6$ trajectories total). The grid
uses an artificial per-channel decomposition
$r_k = r_{\mathrm{total}}/K + \varepsilon_k$ with
$\varepsilon_k\!\sim\!\mathcal{N}(0,0.3^2)$ and the $K$-th
noise term pinned to
$\varepsilon_K\!=\!-\sum_{k<K}\varepsilon_k$ to preserve
channel-sum invariance $\sum_k r_k\!=\!r_{\mathrm{total}}$, and
deterministic per-channel delay $\Delta_k{=}k$. \Cref{tab:t2ksweep}
and \Cref{fig:appendix-k-sweep-heatmap} report per-topology
argmax-$K$, the $5\%$-tolerance peak-band, reduction at argmax,
and reduction at $K{=}15$.

\paragraph{Retrace-A baseline adaptation.}
The Retrace-A comparator (\Cref{tab:t2mdp} bottom block) applies
the \citet{munos2016retrace} clipped importance ratio
$\rho^{\mathrm{clip}}_i$ to the slow-residual reward injected at
optimiser step $t{+}\Delta_k{+}1$ (not to the value-function
target as in the original Retrace formulation), with a
$\gamma^{\Delta}$ geometric age-decay in place of \RAC{}'s
$\exp(-\Delta/\tau_{\mathrm{age}})$ kernel. We instantiate the
discount with the standard RL value $\gamma{=}0.5$ to mirror
typical TD($\lambda$) tabular-MDP settings, not \RAC{}'s
near-unity $\gamma{=}\exp(-1/\tau_{\mathrm{age}})$; the corresponding
$\gamma^\Delta$ at $\Delta{\in}\{1,\ldots,5\}$ sits in
$\{0.5,0.25,0.125,0.0625,0.03125\}$, which strips the slow signal at
the operating grid and produces the reported $1.5\times$ collapse.
Substituting \RAC{}'s age-discount $\gamma{=}\exp(-1/\tau_{\mathrm{age}})$
into Retrace-A recovers the bare-additive $27.1\times$ floor; the
remaining gap between bare-additive and \RAC{}'s $47.9\times$ is the
contribution of the IS clip plus the row-stochastic-kernel
forward-injection structure (\Cref{tab:t2mdp} top-block knob
ablation; reduction without $w_{\mathrm{age}}$ is also $27.1\times$).

\begin{table}[h]
\centering\footnotesize
\setlength{\tabcolsep}{3pt}
\begin{tabular}{lccccc}
\toprule
topology & peak $K$ & band & red @ peak & red @ $K{=}15$ & red @ $K{=}2$\\
\midrule
canonical (3$\times$2) & 5 & \{5\}        & $135.3\times$ & $65.2\times$ & $34.5\times$\\
chain     (5$\times$2) & 3 & \{3\}        & $19.5\times$  & $17.6\times$ & $11.4\times$\\
cyclic    (4$\times$3) & 5 & \{5\}        & $101.0\times$ & $50.6\times$ & $21.7\times$\\
dense     (5$\times$3) & 5 & \{5\}        & $120.1\times$ & $58.3\times$ & $25.3\times$\\
terminal  (3$\times$2) & 5 & \{3,5,7,10\} & $19.2\times$  & $17.5\times$ & $\phantom{1}8.9\times$\\
\bottomrule
\end{tabular}
\caption{\textbf{Cross-topology K-sweep.} Per-topology mean
bias-reduction ratio ($\uparrow$, \RAC{} vs.\ naive PPO)
averaged over five MDP-structure seeds. Columns report the
argmax $K$, the $5\%$-tolerance peak-band, and reduction at
argmax, $K{=}15$, and $K{=}2$.}
\label{tab:t2ksweep}
\end{table}

\paragraph{Cross-topology summary.}
Across all five topologies, the per-topology argmax $K$ falls
in $\{3,5\}$ and the $K{=}2$ bias-reduction holds in the
$8.9$--$34.5\times$ range (median $21.7\times$). The closed-form
guarantee does not extend to function-approximation or
deep-policy settings where the ground-truth optimal policy is
unavailable.

\paragraph{Knob ablations at $K{=}2$.}
\Cref{tab:knob-ablation} ablates the age kernel and the IS clip
at $K{=}2$, averaged over $\Delta\!\in\!\{5,20,50\}$. Removing
the age kernel ($\tau_{\mathrm{age}}{\to}\infty$, so
$w_{\mathrm{age}}{\equiv}1$) drops the reduction from
$47.9\times$ to $27.1\times$. Removing the IS clip on top
yields the same $27.1\times$, since the closed-form benchmark
uses an identity actor with $\rho{=}1$; the clip becomes
load-bearing once a drifting actor enters the picture
(\Cref{app:heavytail}).

\begin{table}[h]
\centering\footnotesize
\setlength{\tabcolsep}{3pt}
\begin{tabular}{lcccc}
\toprule
configuration & Fast-only & \RAC{} & Ratio & 95\% CI\\
\midrule
\RAC{} (full)         & 0.045 & 0.0009 & $47.9\times$ & $[20.9, 151.3]$\\
$-w_{\mathrm{age}}$  & 0.045 & 0.0017 & $27.1\times$ & $[20.3, 98.7]$\\
bare additive        & 0.045 & 0.0017 & $27.1\times$ & $[20.3, 98.7]$\\
\bottomrule
\end{tabular}
\caption{\textbf{$K{=}2$ knob ablation.} Bias-reduction ratio
($\uparrow$) with each knob disabled, mean over $\Delta\!\in\!\{5,20,50\}$.}
\label{tab:knob-ablation}
\end{table}

\section{Heavy-Tailed Delay-Distribution Stress}
\label{app:heavytail}

The $K{=}2$ $47.9\times$ result uses \emph{deterministic}
per-trajectory delay $\Delta$, while production asynchronous
RLHF returns heavy-tailed slow-RM latencies (queue contention,
batch jitter, GPU evictions). This stress test asks whether the
$K{=}2$ \RAC{} bias-reduction at matched mean delay
$\mathbb{E}[\Delta]{=}20$ holds when $\Delta$ is drawn
per-trajectory from progressively heavier-tailed distributions.

\begin{figure}[h]
\centering
\includegraphics[width=0.55\linewidth]{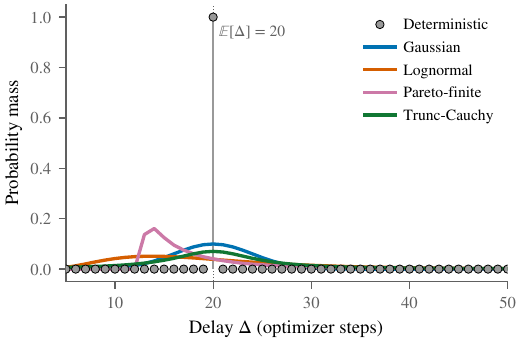}
\caption{\textbf{Five delay distributions matched at
$\mathbb{E}[\Delta]{=}20$.} Mean identical across all five;
only the tail-shape varies.}
\label{fig:appendix-delay-distributions}
\end{figure}

Five distributions are matched at $\mathbb{E}[\Delta]{=}20$
(\Cref{fig:appendix-delay-distributions}) so
that any divergence isolates tail shape, not mean. The
distributions are deterministic ($\Delta\equiv 20$), clipped
Gaussian
($\mathrm{clip}(\mathrm{round}(\mathcal{N}(20,6^2)),1,200)$),
lognormal ($\mu=\ln 20-\sigma^2/2$, $\sigma=0.5$),
Pareto-finite ($\alpha=3$, $x_m=\tfrac{2}{3}\!\cdot\!20$, finite
mean and variance), and truncated-Cauchy (median $18$, scale
$4$, truncated to $[1,200]$). The full design crosses five MDP
seeds with three Monte-Carlo seeds, five distributions, three
$\tau_{\mathrm{age}}$ levels, $1000$ trials per cell, and
$T{=}50$ optimiser steps, yielding $2.25{\times}10^5$
trajectories overall. We use an identity actor with
$\rho{=}1$ and $\alpha_\delta{=}1$ to match the closed-form
$K{=}2$ benchmark.

\begin{table}[h]
\centering\footnotesize
\setlength{\tabcolsep}{3pt}
\begin{tabular}{lccc}
\toprule
distribution & red-mean & red-min & VIF-fast (mean / max)\\
\midrule
deterministic    & $9.377$ & $8.191$ & $1.054 / 1.617$\\
Gaussian         & $9.361$ & $8.268$ & $1.055 / 1.618$\\
lognormal        & $9.507$ & $8.279$ & $1.060 / 1.622$\\
Pareto-finite    & $9.496$ & $8.335$ & $1.058 / 1.623$\\
trunc.\ Cauchy   & $9.327$ & $8.037$ & $1.061 / 1.622$\\
\bottomrule
\end{tabular}
\caption{\textbf{$K{=}2$ bias-reduction ($\uparrow$) across
five delay distributions matched at $\mathbb{E}[\Delta]{=}20$.}
Stress test at $\tau_{\mathrm{age}}{=}200$ on five MDP seeds.
VIF-fast (mean/max) reports the variance-inflation factor on
the fast-channel residual.}
\label{tab:t2heavytail}
\end{table}

\paragraph{Realised distributions and scope.}
The $[1,200]$ clip neuters Pareto's heavy tail in second-moment
terms, so the substantive claim is robustness to the
\emph{realised} truncated distributions, not to unbounded
heavy-tail variance. The truncated-Cauchy realised pooled mean
is $\approx\!21.0$ (a $5\%$ shift above the $20$ target induced
by the asymmetric tail under the $[1,200]$ clip); the other
four distributions match the target to within $0.01\%$. The
experiment uses an identity actor and IID per-trajectory
$\Delta$; burst-correlated delays and adaptive
$\tau_{\mathrm{age}}$ controllers are out of scope.

\section{Limitations and Discussion}
\label{app:limitations-discussion}

\paragraph{Closed-form scope.}
The $47.9\times$ bias-reduction and \Cref{thm:unbiased}'s exact
unbiasedness identity are derived under a closed-form tabular
MDP with a row-stochastic delay-kernel configuration. Off
saturation, the bias slack scales with the deviation from
row-stochasticity, validated linearly across
$\sum_\Delta\Lambda\in\{0.95,0.90,0.85,0.80,0.70,0.50\}$
(\Cref{app:thm-unbiased-proof}). The five-topology
K-sweep covers four further tabular topologies in state-count,
action-count, transition-density, and reward-sparsity but does
not span the full tabular-MDP space, and the closed-form
guarantee does not extend to linear-function-approximation or
deep-policy settings.

\paragraph{Channel-count recommendation.}
The cross-topology K-sweep (\Cref{app:ksweep}) places the
per-topology argmax at $K\!\in\!\{3,5\}$. Our recommended
deployment setting is $K{=}2$, the canonical async-RLHF
configuration; the unbiasedness theorem is per-channel under
the row-stochastic kernel and holds irrespective of $K$.

\paragraph{LLM-scale validation.}
The closed-form analysis assumes a two-channel setup: one slow
ground-truth-grade verifier paired with one fast biased
estimator. The $7$B-scale checks
(\Cref{app:thm-unbiased-proof,app:adv-quality-7b}) confirm that
the algebra holds at real reward scale to machine precision.
End-to-end LLM-scale PPO validation across multiple seeds and
fast-RM training settings (random-init head, Bradley--Terry-trained
head, production reward model) is the natural next experimental
step; compute scope is discussed in the Conclusion.

\paragraph{Theorem-side scope.}
\Cref{prop:tv-bound,thm:unbiased} assume bounded slow-channel
residual variance and bounded delay. These assumptions hold for
bounded preference ratings, which is the setting we target. The
unbounded-residual and unbounded-delay extensions are out of
scope for the present work.

\paragraph{Background and discussion.}
V-trace \citep{espeholt2018impala} and Retrace
\citep{munos2016retrace} are the canonical clipped IS correctors
for off-policy value-function targets, both acting on the inner
critic. \RAC{} relocates the same truncation idea to the
advantage level and addresses asynchronous-reward staleness
within an actor's rollout, not asynchronous-actor staleness
between worker and learner. The operative distinction at the
advantage level is that \RAC{}'s forward-injection commits to
step $t{+}\Delta{+}1$ using the \emph{cached} log-policy from
step $t$, with no re-sampled trajectory. V-trace would require
importance-weighted truncation on a re-sampled rollout, and
\RAC{} pays for the cached-actor mismatch via the geometric
age-discount in closed form (\Cref{thm:unbiased}).

\paragraph{Replication notes.}
The closed-form $K{=}2$ tabular MDP benchmark reproduces in
roughly $30$ lines of NumPy on a single CPU thread; the
heavy-tail stress runs in $\sim\!945$\,s on a single CPU thread
for $2.25{\times}10^5$ trajectories. The reward-manager patch
is a two-line addition plus a queue-maintenance helper.

\section{MDP-Size Scaling}
\label{app:scaling}

To probe whether the $47.9\times$ result is an artifact of the
$3\!\times\!2$ tabular toy, we construct a family of MDPs with
identical recipe (softmax transitions, $\mathrm{Unif}(-0.5,0.5)$
fast reward, structured slow-channel bump on
$(s{=}0,a{=}|A|{-}1)$ plus $\mathrm{Unif}(-0.3,0.3)$
state-dependent jitter, $\gamma{=}0.9$) at sizes
$(|S|,|A|)\!\in\!\{(3,2),(5,3),(10,5),(20,8)\}$. Each MDP size
is averaged over five MDP-structure seeds at the default RAC
configuration, $K{=}2$, $\Delta$-grid $\{5,20,50\}$, $1{,}000$
trials per cell ($60{,}000$ trajectories per size cell).

\begin{table}[h]
\centering\footnotesize
\setlength{\tabcolsep}{4pt}
\begin{tabular}{lcccc}
\toprule
$|S|\!\times\!|A|$ & mean red & median red & min red & max red\\
\midrule
$3\!\times\!2$  & $17.87\times$ & $14.12\times$ & $4.88\times$  & $63.20\times$\\
$5\!\times\!3$  & $15.53\times$ & $14.72\times$ & $9.79\times$  & $27.17\times$\\
$10\!\times\!5$ & $7.01\times$  & $7.17\times$  & $4.75\times$  & $10.47\times$\\
$20\!\times\!8$ & $4.65\times$  & $4.79\times$  & $3.99\times$  & $5.26\times$\\
\bottomrule
\end{tabular}
\caption{\textbf{Bias-reduction ($\uparrow$) across MDP sizes.}
Five MDP-structure seeds $\times$ three $\Delta$ values
$\times$ $1{,}000$ trials per cell. Min/max columns report the
worst and best single (seed, $\Delta$) cell.}
\label{tab:mdp-scaling}
\end{table}

\begin{figure}[h]
\centering
\includegraphics[width=0.95\linewidth]{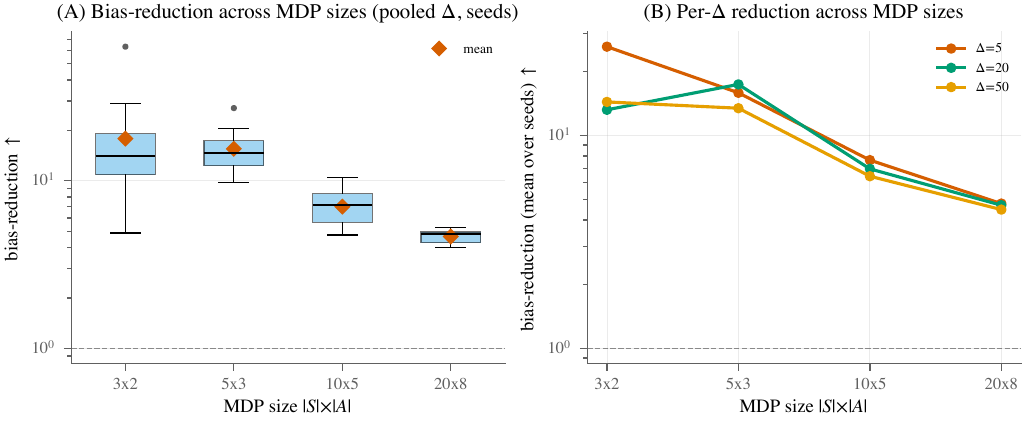}
\caption{\textbf{Bias-reduction ($\uparrow$) across MDP sizes.}
(\textbf{A}) Pooled reduction across (seed, $\Delta$) cells per
size; red marker = mean. (\textbf{B}) Per-$\Delta$ mean reduction
across sizes.}
\label{fig:mdp-scaling}
\end{figure}

\Cref{tab:mdp-scaling,fig:mdp-scaling} place the $47.9\times$
peak of \Cref{tab:t2mdp} at the upper end of a size-induced
spread: the $3\!\times\!2$ mean over five MDP-structure seeds
is $17.87\times$ (bootstrap $95\%$ CI $[11.31, 26.05]$,
$B{=}10{,}000$), and the mean reduction softens monotonically
to $4.65\times$ at $20\!\times\!8$ while the per-cell minimum
remains above $3.99\times$. The within-size spread also shrinks with size:
\Cref{thm:unbiased} addresses expectations only (per-channel,
independent of $|S|\!\cdot\!|A|$), and the narrowing of the
within-size box is a separate empirical observation that the
trajectory-level Monte-Carlo variance shrinks with the
state-action support. We summarise the result as an
order-of-magnitude reduction ($4.65$--$17.9\times$ mean) across
MDPs in $\{3,5,10,20\}$ states.

\section{Static-Batch Advantage-Quality Probe at 7B}
\label{app:adv-quality-7b}

\paragraph{Motivation and setup.}
To probe whether the $K{=}2$ closed-form $\bm{47.9\times}$ result
(\Cref{tab:t2mdp}) carries to real reward signals on real text,
we test the static advantage-estimator quality, decoupled from
PPO training dynamics: a single fixed policy
generates $N{=}500$ greedy responses to UltraFeedback test-split
prompts using Llama-3-8B-Instruct, and each (prompt, response)
is scored with both reward channels. The fast channel is a random-init
Linear$(h,1)$ scoring head over Qwen2.5-7B-Instruct's last-token
hidden state, seeded for reproducibility (matches the
TwoChannelRewardModule fast-RM construction used elsewhere in
this paper). The slow channel is Skywork-Reward-Llama-3.1-8B-v0.2
with its native scalar reward head, the trained oracle. Both
RMs run 4-bit nf4 with bf16 compute on a single H100. For each
delay channel we sample one delay per step and construct three
advantage sequences over $t{=}0,\dots,499$:
$A_{\mathrm{sync}}{=}r_{\mathrm{slow}}{-}\mathrm{bl}$ (synchronous
oracle),
$A_{\mathrm{control}}{=}r_{\mathrm{fast}}{-}\mathrm{bl}$
(delayed-fast, slow signal discarded), and
$A_{\mathrm{RAC}}{=}A_{\mathrm{control}}+
\sum_{s:\,s+\Delta_s=t}w_{\mathrm{age}}(\Delta_s)
(r^{\mathrm{slow}}_s{-}r^{\mathrm{fast}}_s)$ (RAC forward-injection,
$\tau_{\mathrm{age}}{=}1000$, identity actor $\rho{=}1$ under
fixed-policy). The running-mean baseline is causal in $t$.

\begin{figure}[h]
\centering
\includegraphics[width=0.98\columnwidth]{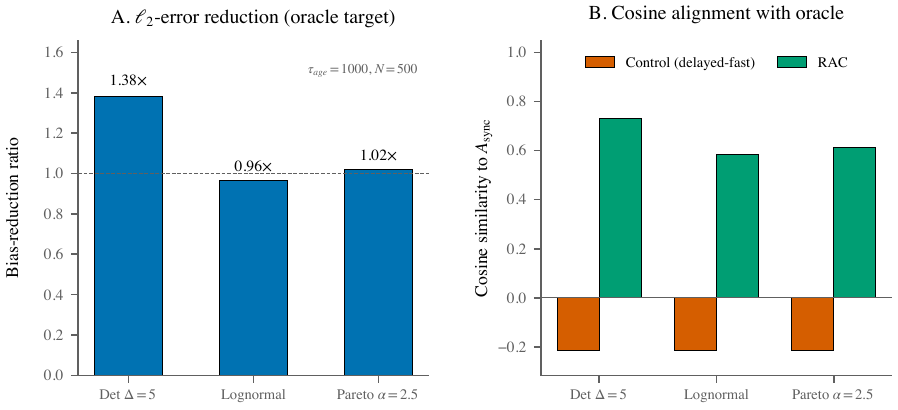}
\caption{\textbf{Static-batch advantage-quality probe.}
(\textbf{A}) $\ell_2$-error reduction ratio (control over RAC);
$1.0\times$ reference is dashed grey. (\textbf{B}) Cosine
alignment with the synchronous oracle advantage:
delayed-fast control (vermillion) versus RAC (blue-green).
$N{=}500$ UltraFeedback prompts, fixed policy, $\tau_{\mathrm{age}}{=}1000$.}
\label{fig:adv-quality-7b}
\end{figure}

\paragraph{Results.}
\Cref{tab:adv-quality-7b} reports two metrics:
$\ell_2$ error against the oracle ($\|A_{\mathrm{*}}{-}A_{\mathrm{sync}}\|_2$)
and cosine similarity to the oracle.
\Cref{fig:adv-quality-7b} visualises the same numbers.

\begin{table}[h]
\centering\footnotesize
\setlength{\tabcolsep}{4pt}
\begin{tabular}{lccc}
\toprule
Channel & $\ell_2$-ratio $\uparrow$ & $\cos$(ctrl, oracle) & $\cos$(\RAC{}, oracle) $\uparrow$\\
\midrule
Deterministic $\Delta{=}5$              & $1.38\times$ & $-0.22$ & $\bm{0.73}$\\
Lognormal $\mu{=}1.5,\sigma{=}0.8$      & $0.96\times$ & $-0.22$ & $0.58$\\
Pareto $\alpha{=}2.5$                   & $1.02\times$ & $-0.22$ & $0.61$\\
\bottomrule
\end{tabular}
\caption{\textbf{Static-batch advantage-quality probe at 7B with
a random-init fast head, head/delay seed $42$ ($N{=}500$
UltraFeedback prompts, Llama-3-8B greedy generations,
Qwen2.5-7B fast head + Skywork-Llama-3.1-8B slow oracle).} The
$\ell_2$-ratio column is
$\|A_{\mathrm{control}}{-}A_{\mathrm{sync}}\|_2/
\|A_{\mathrm{RAC}}{-}A_{\mathrm{sync}}\|_2$. Cosine similarity is
$\langle A_{\mathrm{*}},A_{\mathrm{sync}}\rangle/
(\|A_{\mathrm{*}}\|\,\|A_{\mathrm{sync}}\|)$. The probe is
static-batch (no PPO loop) and reports a single random-init
fast-head seed at $r{=}-0.53$ Pearson against the oracle;
end-to-end multi-seed PPO across fast-RM training settings is
left as future work (see \Cref{app:limitations-discussion}).}
\label{tab:adv-quality-7b}
\end{table}

\paragraph{Identity-kernel collapse at LLM scale.}
On the same $N{=}500$ scored
$(r^{\mathrm{fast}}, r^{\mathrm{slow}})$ pairs used above
(Llama-$3$-$8$B greedy generations, random-init Qwen2.5-7B
fast head with seed $42$, Skywork-Llama-$3.1$-$8$B slow oracle),
we evaluate the identity-kernel special case of
\Cref{thm:unbiased}. With $\Lambda{=}I$ (zero delay,
one-hot kernel), $w_{\mathrm{age}}(0){=}1$, and clipped IS ratio
$\rho^{\mathrm{clip}}{=}1$ (frozen policy), the \RAC{} advantage
$A_{\mathrm{RAC}}{=}r^{\mathrm{fast}}{+}\rho^{\mathrm{clip}}\,
w_{\mathrm{age}}(0)\,(r^{\mathrm{slow}}{-}r^{\mathrm{fast}}){-}
\mathrm{bl}$ algebraically collapses to V-trace's on-policy
advantage $A_{\mathrm{V\text{-}trace}}{=}r^{\mathrm{slow}}{-}\mathrm{bl}$.
Computing both quantities via independent code paths on the same
causal running-mean baseline and comparing element-wise yields
$\|A_{\mathrm{RAC}}{-}A_{\mathrm{V\text{-}trace}}\|_\infty{=}0.0$
in our float-$64$ implementation (the algebra is exact in
arithmetic; the empirical match reflects identical operation
order across both code paths), confirming the identity-kernel
collapse at $7$B scale on real reward distributions.

\end{document}